\documentclass[journal]{IEEEtran}
\usepackage[utf8]{inputenc}
\usepackage{fullpage}
\usepackage{amsmath}
\usepackage{amsfonts}
\usepackage{amssymb}
\usepackage[pdftex]{hyperref}
\usepackage{graphicx}
\usepackage{algorithmic}
\usepackage{algorithm}
\usepackage{color}
\usepackage[numbers,sort&compress]{natbib}
\usepackage{authblk}
\usepackage{subfigure}
\usepackage{multirow}
\usepackage{booktabs}
\usepackage{tikz}
\usepackage[raggedright]{sidecap}
\usepackage{balance}

\definecolor{darkgreen}{rgb}{0,0.6,0.2}

\title{Temporal Overdrive Recurrent Neural Network}

\author[1]{Filippo Maria Bianchi\thanks{filippo.m.bianchi@uit.no}\thanks{Corresponding author}}
\author[1]{Michael Kampffmeyer\thanks{michael.c.kampffmeyer@uit.no}}
\author[2]{Enrico Maiorino\thanks{enrico.maiorino@uniroma1.it}}
\author[1]{Robert Jenssen\thanks{robert.jenssen@uit.no}}
\affil[1]{Machine Learning Group, Dept. of Physics and Technology, University of Troms\o{}, Norway}
\affil[2]{Dept. of Information, Electronic and Communication Engineering, Sapienza University, Rome, Italy}

\providecommand{\keywords}[1]{\textbf{\textit{Keywords---}} #1}
\begin{document}

\maketitle

\begin{abstract}
In this work we present a novel recurrent neural network architecture designed to model systems characterized by multiple characteristic timescales in their dynamics. 
The proposed network is composed by several recurrent groups of neurons that are trained to separately adapt to each timescale, in order to improve the system identification process.
We test our framework on time series prediction tasks and we show some promising, preliminary results achieved on synthetic data. To evaluate the capabilities of our network, we compare the performance with several state-of-the-art recurrent architectures.\\
\end{abstract}

%%%%%%%%%%%%%%%%%%%%%%%%%%%%%%%%%%%%%%%%%%%%%%%%%%%%%%%%
%%%%%%%%%%%%%%%%%%%%% INTRODUCTION %%%%%%%%%%%%%%%%%%%%%
%%%%%%%%%%%%%%%%%%%%%%%%%%%%%%%%%%%%%%%%%%%%%%%%%%%%%%%%
\section{Introduction}

% motivation
Time series analysis focuses on reconstructing the properties of a dynamical system from a sequence of values, produced from a noisy measurement process acting on the state space of the system. 
Correct modeling is necessary to understand, characterize the dynamics of the system and, consequently, to predict its future behavior. A Recurrent Neural Network (RNN) is an universal approximator of Lebesgue measurable dynamical systems and is a powerful tool for predicting time series \cite{Hammer20041061}. 
In its basic formulation, the whole state of a RNN evolves according to precise timing. However, in order to model a system whose internal dynamic variables evolve at different timescales, a RNN is expected to learn both short and long-term dependencies. 
The capability of modeling multiple time scales is particularly important in the prediction of real-world time series, since they are generally characterized by multiple seasonalities \cite{Gould2008207}.

% Proposed framework
In this paper we propose a novel RNN architecture, specifically designed for modeling a system characterized by heterogeneous dynamics operating at different timescales. 
We combine the idea of using bandpass neurons in a randomly connected recurrent layer, an inherent feature of reservoir computing approaches \cite{Jaeger2007335}, with the gradient-based optimization, adopted in classic RNN architectures. 
Contrarily to standard RNNs, where the internal state of the neurons is influenced by all the frequencies of the input signal, in our network specific groups of neurons update their states according to different frequencies in the input signal.
Therefore, since the hidden neurons can operate at temporal resolutions that are slower than the fastest frequencies in the input, we named our network Temporal Overdrive RNN (TORNN), after the mechanism in vehicles engine that, under certain conditions, decouples the speed of the wheels from the revolutions of the engine.

TORNN is capable of modeling with high accuracy the timescales in the dynamics of the underlying system. 
In the proposed architecture, the hidden recurrent layer is partitioned in different groups, each one specialized in modeling dynamics at a single timescale. 
This is achieved by organizing the recurrent layer in groups of bandpass neurons with shared parameters, which determine the frequency band to operate on. 
During training, while the neuron connection weights are fixed to their initial random values, the bandpass parameters are learned via backpropagation. 
In this way, the recurrent layer dynamics are smoothed out by the bandpass filtering in order to adapt to the relevant timescale. 
The main hyperparameter required by TORNN is the number of neuron groups composing the recurrent layer, that has to be chosen by considering the expected number of separate timescales characterizing the analyzed process. 
This information can be easily retrieved through an analysis in the frequency domain and/or by \textit{a priori} knowledge of the input data.
The network also depends on few other structural hyperparameters, which are not critical as the final performance of the model is not particularly sensitive to their tuning. 
Since the pair of bandpass parameters are the only weights trained in each group of neurons, the total number of parameters to be learned are considerably less than in other RNN architectures, with a consequent simplification of the training procedure.

% paper organization
In this work we focus on the prediction of real-valued time series, generated from a system whose internal dynamics operate at different timescales. 
In Sect. \ref{sec:related_works} we review previous works that dealt with the problem of modeling dependencies across different time scales in the data with RNNs. 
In Sect. \ref{sec:model} we provide the details of the proposed architecture. 
In Sect. \ref{sec:experiments} we present some initial results on synthetic data obtained by the proposed architecture and other state-of-the-art alternatives. 
Finally, in Sect. \ref{sec:conclusions} we draw our conclusions.

%%%%%%%%%%%%%%%%%%%%%%%%%%%%%%%%%%%%%%%%%%%%%%%%%%%%%%%%%
%%%%%%%%%%%%%%%%%%%%% RELATED WORKS %%%%%%%%%%%%%%%%%%%%%
%%%%%%%%%%%%%%%%%%%%%%%%%%%%%%%%%%%%%%%%%%%%%%%%%%%%%%%%%
\section{Related works}
\label{sec:related_works}

% Long term dependencies
Classic RNN architectures struggle to model different temporal resolutions, especially in presence of long-term relations that are poorly handled, due to the issue of the vanishing gradient \cite{279181}.
Several solutions have been proposed in the past to deal with long-term dependencies \cite{lin1996learning} and an important milestone has been reached with the introduction of Long Short Term Memory (LSTM) \cite{hochreiter1997long}. 
This architecture has been shown to be robust to the vanishing gradient problem, thanks to its gated access mechanism to its neurons' state.
This results in the LSTM networks' capability to handle long time lags between events and to automatically adapt to multiple time granularity.
In particular, the LSTM neurons, referred to as ``cells'', are composed by an internal state that can be accessed through three gates: the input gate, that controls whether the current input value should replace the existing state; the forget gate, that controls the reset of the cell's state; the output gate, that determines whether the unit should output its current state.
An additional component in several implementations of LSTM cells are the peephole connections, which allow gate layers to look directly at the cell state.
A recent variant of the traditional LSTM architecture is the Gated Recurrent Unit (GRU) \cite{cho2014learning}, which always exposes its internal state completely, without implementing mechanisms to control the degree of exposure.
In GRU, both peephole connections and output activation functions are removed, while the input and the forget gate are coupled into an update gate. 
While LSTM and GRU excel in learning at the same time long and short time relationships, they are difficult to train due to an objective function landscape characterized by a high dimensionality and several saddle points \cite{glorot2010understanding}. 
Additionally, these networks depend on several hyperparameters, whose tuning is non-trivial, and the utility of their various computational components depends on the application at hand \cite{DBLP:journals/corr/GreffSKSS15}.

% Layered networks
An initial attempt to model multiple dynamics and timescales was based on the idea that temporal relationships are structured hierarchically, hence the RNN should be organized accordingly \cite{el1995hierarchical}. 
The resulting architecture managed to improve the modeling of slow-changing contexts. 
An analogous hierarchical organization has been implemented by stacking multiple recurrent layers \cite{graves2013generating}. 
In the same spirit, a more complex stacked architecture called Gated Feedback Recurrent Neural Networks has been proposed in Ref. \cite{chung2015gated}. 
In this architecture, the recurrent layers are connected by gated-feedback connections which allow them to operate at different timescales.
By referring to the unfolding in time of the recurrent network, the states of consecutive time steps are fully connected and the strength of the connections is trained by gradient descent. 
The main shortcoming of these layered architectures is the high amount of parameters that must be learned in order to adapt the network to process the right time scales. 
This results in a long training time, with the possibility of overfitting the training data.

The issue of modeling multiple timescales has also been discussed within the framework of reservoir computing, a quite recent approach to temporal signal processing that leverages on a large, randomly connected and untrained recurrent layer called reservoir. 
The desired output is computed by a linear memory-less readout, which maps the instantaneous states of the network and is usually trained by linear regression. 
The original reservoir computing architectures, known as Echo State Networks \cite{jaeger2001echo} and Liquid State Machines \cite{RIS_0}, are characterized by a fading memory that prevents to model slow periodicities that extend beyond the memory capacity of the reservoir. 
A solution proposed in Ref. \cite{Jaeger2007335} is to slow down the dynamics of the reservoir by using leaky integrator units as neurons. 
These units have individual state dynamics that can model different temporal granularity in the target task. 
Leaky integrators behave as lowpass filters, whose cutoff frequency is controlled by the leakage parameter. 
Precise timing phenomena emerge from the dynamics of a random connected network implementing these processing units, without the requirement of dedicated timing mechanism, such as clocks or tapped delay lines.
A different kind of unit that implements a bandpass filter has been proposed to encourage the decoupling of the dynamics within a single reservoir and to create richer and more diverse echoes, which increase the processing power of the reservoir \cite{siewert2007echo}. 
Improved results have been achieved by means of neurons implementing more advanced digital bandpass filters, with particular frequency-domain characteristics \cite{4634252}.
Differently from other strategies where the optimal timescales for the task at hand are learned from data, reservoir computing follows a ``brute force'' approach. 
In fact, not only a large, sparse and randomly connected reservoir is generated in order to provide rich and heterogeneous dynamics, but also a high amount of filters are initialized with random cutoff frequencies, to encode a wide range of timescales in the recurrent layer. 
By providing filters that cover the whole spectrum, the correct timings required to model the target dynamics are likely to be provided, at the cost of a considerably redundant representation of the system.
An alternative approach has also been proposed, where the filters are tuned manually according to the information of the frequency spectrum of the desired response signal \cite{4634252}.

%%%%%%%%%%%%%%%%%%%%%%%%%%%%%%%%%%%%%%%%%%%%%%%%%%%%%%%%%%%%%%
%%%%%%%%%%%%%%%%%%%%% MODEL ARCHITECTURE %%%%%%%%%%%%%%%%%%%%%
%%%%%%%%%%%%%%%%%%%%%%%%%%%%%%%%%%%%%%%%%%%%%%%%%%%%%%%%%%%%%%
\section{Model architecture}
\label{sec:model}

In this section we discuss the details of TORNN architecture. 
Let us consider a time series $\boldsymbol{x} = \{ x[t] \}_{t=1}^{T}$, whose values are relative to the measurement of the states of a system, characterized by slow and fast dynamics. 
Each dynamical component is modeled by a specialized group of neurons in the recurrent layer which operates at the same characteristic frequencies. 
The number of groups $K$ is equal to the number of main dynamical components of the system and can be determined through a frequency analysis on the input signal $\boldsymbol{x}$.
In this work, by means of a power spectral density estimate, we identify $K$ as the number of peaks in the power spectrum, but alternative approaches are possible.

In order to operate only on the portion of the spectrum located around one of the maxima, each processing unit implements a bandpass filter, whose configuration is identical to the one of the other units in the same group. 
As it is shown later in Sect. \ref{sec:band_pass}, the transfer function of each bandpass filter depends on two parameters related to the two cutoff frequencies, but also on the connection weights of input and recurrent layers. 
If the configuration of these connections is modified, the frequency response of each filter will change as well.
In TORNN we implement an hybrid approach. Analogously to reservoir computing methodologies, input and recurrent connections are randomly initialized and kept unchanged, while the pair of parameters that define the bandpass in each group of neurons are trained via backpropagation, along with the output layer.

In the following, we first explain how the recurrent layer is structured. Then, we describe how the bandpass filters are implemented in the hidden processing units and, finally, we discuss the learning procedure adopted to train the parameters in the model.

% ------------- Recurrent layer structure -------------
\subsection{Recurrent layer structure}

The recurrent layer of the network is randomly generated, under the constraint that the connections form a structure with $K$ groups. 
Each group $\mathcal{C}_k$ contains $N$ neurons, which are strongly connected to the other neurons in  $\mathcal{C}_k$. 
Additionally, in order to allow for some weak coupling between different dynamical components in the modeled system, we also generate a small amount of connections between neurons belonging to different groups. 
This allows synchronization mechanisms between different groups, which are required if the different dynamics of the modeled system are coupled. 
For intra-group connections, we define a probability $p$ that determines the presence of a link $e_{i,j}$ between the neurons $n_i$ and $n_j$ of the same group $\mathcal{C}_k$. 
A second probability $q$ determines the presence of an inter-group connection $e_{i,j}$ between the neurons $n_i$ and $n_j$ of different groups.
To guarantee higher intra-group connectivity we set $p,q$ such that $p \gg q$. 

\begin{figure}
	\centering
		\includegraphics[width=0.3\textwidth]{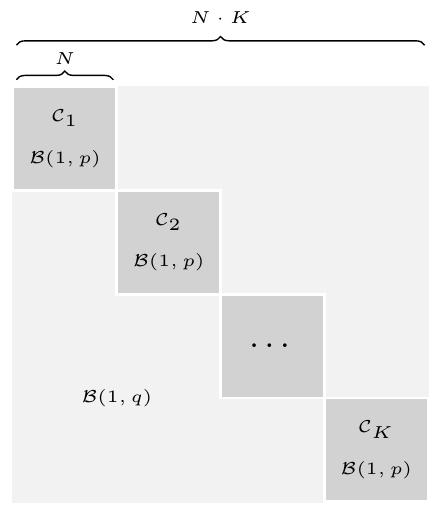}
	\caption{Structure of the recurrent layer, encoded in the weight matrix $\mathbf{W}_r$. Inter-group connections are drawn from a binomial distribution $\mathcal{B}(1,p)$, while intra-group connections are drawn from $\mathcal{B}(1,q)$. To ensure inter-group connections to be more dense than intra-group connections, we set $p \gg q$.}
	\label{fig:Rec_struct}
\end{figure}

To define the recurrent layer connections, we first generate a Boolean squared matrix $\mathbf{W}_r \in \mathbb{R}^{NK \times NK}$ by drawing values from a binomial distribution $\mathcal{B}(1,p)$ for elements belonging to one of $K$ blocks on the diagonal, or a binomial distribution $\mathcal{B}(1,q)$, for the remaining elements. 
The structure of $\mathbf{W}_r$ is depicted in Fig. \ref{fig:Rec_struct}.
To obtain a higher degree of heterogeneity among neuron activations and to enrich internal dynamics \cite{ozturk2007analysis}, each non-zero connection $e_{i,j}$ is multiplied by a value, drawn from a normal distribution $\mathcal{N}(0,1)$.
Finally, to guarantee stability in the network \cite{bianchi2016investigating}, we rescale the spectral radius of $\mathbf{W}_r$ to a value lower than 1.

Analogously to the connections in the recurrent layer, the input weights (stored in the matrix $\mathbf{W}_i \in \mathbb{R}^{I \times NK}$, with $I$ the input size) are randomly initialized with values drawn from a distribution $\mathcal{N}(0,1)$ and are kept untrained as well.

% ------------- bandpass units -------------
\subsection{bandpass processing units}
\label{sec:band_pass}

A leaky integrator neuron outputs a weighted average -- controlled by a leakage rate $\gamma$ -- of the new activation of the neuron with the one from the previous time interval \cite{bengio2013advances,Jaeger2007335}.
A leaky neuron acts as a lowpass filter, whose cutoff frequency is proportional to $\gamma$. Its state-update equation is defined as
\begin{equation}
\label{eq:leaky_neuron1}
    x[t+1] = (1-\gamma)x[t] + \gamma f(\mathbf{W}_r x[t] + \mathbf{W}_i u[t+1])
\end{equation}
or, by following an alternative definition \cite{Schrauwen2007}, as
\begin{equation}
\label{eq:leaky_neuron2}
    x[t+1] =  f((1-\gamma)x[t] + \gamma\mathbf{W}_r x[t] + \mathbf{W}_i u[t+1])
\end{equation}
where $f(\cdot)$ is a $tanh$ (or a sigmoid) function. 
If $\gamma = 0$, the new state at time $t+1$ maintains the value of state $t$. If $\gamma=1$, the state-update equation reduces to a classic nonlinear transfer function.

In the integrator of Eq. \ref{eq:leaky_neuron2}, the non-linearity $f(\cdot)$ is applied also to the leakage term. This guarantees stability, since the poles of the transfer function are constrained to the unit circle, and has the advantage of no computational overhead, since the integrators can be incorporated in $\mathbf{W}_r$ \cite{Schrauwen2007}. This integrator always leaks, due to the contracting property of the non-linear mapping of the hyperbolic tangent upon itself. Since this is not an issue in TORNN, these are the filters we chose to implement.

As previously discussed, we want the processing units in each group of the recurrent layer to act as bandpass filters, which allow signals between two specific frequencies to pass, but discriminate against signals at other frequencies. In particular, we want the neurons in the group $\mathcal{C}_k$ to reject the frequency components that are not close to a given peak in the spectrum.
A bandpass filter can be generated by combining a lowpass filter with a highpass filter (implemented by negating the lowpass filter). Alternatively, a bandpass filter is obtained by combining two lowpass filters. In this latter case, according to Eq. \ref{eq:leaky_neuron2}, the state update equation of a bandpass neuron reads as
\begin{equation}
\label{eq:band_pass}
\begin{aligned}
    x^{'}[t+1] &= f((1-\gamma_1) x^{'}[t] +\\
    & + \gamma_1 \mathbf{W}_r x[t] + \mathbf{W}_i u[t+1]), \\
    x^{''}[t+1] &= (1-\gamma_2) x^{''}[t] +  \gamma_2 x^{'}[t+1], \\
    x[t+1] &= x^{'}[t+1] - x^{''}[t+1].
\end{aligned}
\end{equation}
The parameters $\gamma_1$ and $\gamma_2$ control respectively the high and low frequency cutoffs of the bandpass filter. 
Interestingly, the filter has a transfer function equivalent to the one of an analogue electronic RCL circuit \cite{siewert2007echo}.

% ------------- Learning -------------
\subsection{Learning}

The bandpass filters implemented by the units in the group $\mathcal{C}_k$ must specialize to pass only the frequencies in neighborhood of one of the peaks in the power spectrum of $\boldsymbol{x}$. 
According to Eq. \ref{eq:band_pass}, to control the band to be passed by the filters in each group, one must tune the parameters $\gamma_1$ and $\gamma_2$.
However, in practical cases the bandpass width of each filter is not easy to determine in advance, as the neighborhood of each peak in the power spectrum could either include noise or useful information for the problem at hand.
Most importantly, $\gamma_1$ and $\gamma_2$ are related to the effective cutoff frequencies in the spectrum through a highly non-linear dependency and the desired response of the filter is difficult to determine. Therefore, we follow a data-driven approach to automatically learn the control parameters by means of a gradient descent optimization, which accounts the nature of the task at hand.

\begin{figure}[!ht]
	\centering
		\includegraphics[width=0.7\columnwidth]{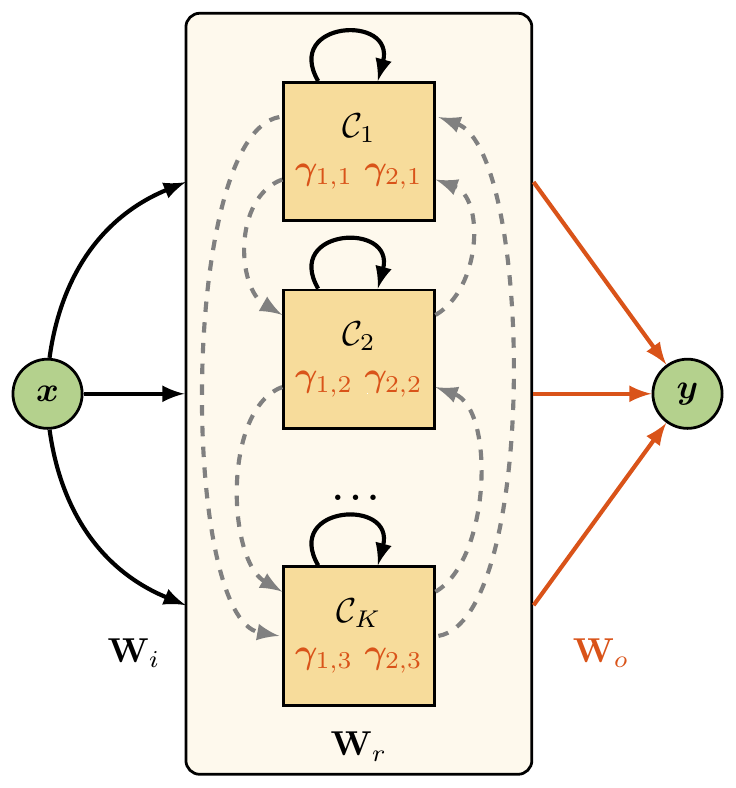}
	\caption{Input connections (encoded in $\mathbf{W}_i$) and recurrent layer connections (encoded in $\mathbf{W}_r$) are initialized pseudo-randomly and they are kept fixed afterwards. 
	In the recurrent layer, the connections between neurons in the same group, depicted as solid black lines, are more dense than the connections between neurons in different groups (dashed gray lines). In red, we marked the elements that are trained by means of a gradient descent optimization, which are the parameters that determine the cutoff frequencies in each group $\mathcal{C}_k$ ($\gamma_{1,k}$ and $\gamma_{2,k}$) and the weights connecting the recurrent layer with the network output ($\mathbf{W}_o$).}
	\label{fig:architecture}
\end{figure}

The last component in the architecture is a dense layer, composed by a set of weights stored in the matrix $\mathbf{W}_o \in \mathbb{R}^{H \times O}$, which combines the output of the neurons in the recurrent layer to produce the $O$-dimensional output.
A schematic depiction of the whole architecture is reported in Fig. \ref{fig:architecture}.

Like the filter parameters, also the connection weights in $\mathbf{W}_o$ are learned through a gradient descent optimization. 
We opted for a standard backpropagation through time (BPTT) parametrized by a threshold $\tau_{\text{tnc}}$, which determines the truncation of the unfolding of the network in time.
To constrain the parameters $\gamma_1$ and $\gamma_2$ to lay within the $[0,1]$ interval (required from Eq. \ref{eq:band_pass}), during the optimization we apply a sigmoid function to the values of $\gamma_1$ and $\gamma_2$ learned by the gradient descent.

%%%%%%%%%%%%%%%%%%%%%%%%%%%%%%%%%%%%%%%%%%%%%%%%%%%%%%%
%%%%%%%%%%%%%%%%%%%%% EXPERIMENTS %%%%%%%%%%%%%%%%%%%%%
%%%%%%%%%%%%%%%%%%%%%%%%%%%%%%%%%%%%%%%%%%%%%%%%%%%%%%%
\section{Experiments}
\label{sec:experiments}

% Motivation of the experiment
In this section, we evaluate the capability of TORNN to model multiple dynamics characterized by different timescales. 
To test our model, we consider the prediction of a time series generated by superimposed sine waves, whose frequencies are incommensurable. 
Since the ratio of the frequencies is an irrational number, the periods of the sine waves do not cross with each other and hence the period of the resulting signal is, in theory, infinite.
This academic, yet important task, has been previously considered to test the memory capacity of a recurrent neural network \cite{jaeger2004harnessing}.
Indeed, to accurately predict the unseen values of the time series, the network would require a huge amount of memory.
Here, we tackle the problem from a different perspective. 
We show that if the network is capable of learning how to model the dynamics of each single oscillator separately, the prediction task can be solved accurately with a limited amount of computational resources.

% Generation of the data
In the following, we try to solve several time series prediction tasks of increasing difficulty. 
Each time series is obtained by superimposing sinusoids, whose frequencies are pairwise incommensurable. 
The sinusoids are generated by multiplying a frequency $\phi$ by distinct integer powers of a transcendental number, such as $e$ (Euler number) or $\pi$. In our experiments we chose $e$ as the base number and the considered time series have the form
\begin{equation}
\label{eq:mso}
x_K[t] = \sum \limits_{k=1}^{K} sin( e^k \phi )[t],
\end{equation}
where $K$ is the number of superimposed sine waves. The difficulty of the prediction task is controlled by increasing the number of components $K$ and by adding to the time series a white noise $n[t] \sim \mathcal{N}(0, 1)$. In the experiments, we set a noise-to-signal ratio of 0.2 and we evaluated time series with $T=5000$ time-steps.

% Setup
To quantify the performance of TORNN on each task, we compare its prediction accuracy with the ones achieved by Elmann-RNN (ERNN), LSTM, GRU and ESN.
To initialize the trainable weights in TORNN, ERNN, LSTM and GRU, we draw their initial values from a Gaussian distribution $\mathcal{N}(0, 1/\sqrt{d})$, $d$ being the number of processing units in the successive layer in the network \cite{glorot2010understanding}. 
For each network we used Adam algorithm as the optimizer of the gradient descent step \cite{kingma2014adam}. 
ERNN, LSTM and GRU are configured to use a comparable number of parameters (approximately 8500). 
For TORNN and ESN instead, we instantiate 20 neurons in the recurrent layer for each sinusoidal component in the signal. 
Therefore, the number of parameters trained in TORNN is $(2 \times K) + (20 \times K) + (1)$, which are, respectively, the filter parameters for all groups, the number of connections from the recurrent layer to the output and the bias.
While the number of neurons in ESN reservoirs is usually much larger, the reason for this experimental setup is to show that TORNN manages to handle the same computational resources of an ESN more effectively.

The details of the configuration in each network are reported in Tab. \ref{tab:config}.  
The model parameters of ESN are not learned through gradient descent optimization. 
Indeed, the output weights in the readout are computed through a simple ridge-regression, an optimization problem that can be expressed in close form and solved in a time that grows as the cube of the number of neurons \cite{bianchi2015prediction}. 
On the other hand, a correct setup of the hyperparameters in the ESN is in general more critical than in other architectures and they are usually tuned through cross validation procedures. 
In our experiments, we used a genetic algorithm to search for the optimal hyperparameters values. 
We followed the same approach described in \cite{DBLP:journals/corr/LokseBJ16}, to which we refer the interested reader for further details. 
The bounds considered in the search of each hyperparameter are reported in Tab \ref{tab:config}.

\bgroup
\def\arraystretch{1} %vertical padding
\setlength\tabcolsep{1em} %horizontal padding
  \begin{table*}[!ht]\scriptsize\centering
  \caption{Summary of the hyperparameters configuration for each network. ERNN, LSTM, GRU: number of processing units ($N_r$), number of trainable parameters (\# params), $\mathrm{L}_2$ regularization of $\mathbf{W}_o$ weights ($\lambda$), gradient truncation in BPPT ($\tau_{tnc}$). ESN (admissible range of hyperparameters, optimized with a genetic algorithm \cite{DBLP:journals/corr/LokseBJ16}): neurons of the reservoir ($N_r$ -- not optimized), spectral radius ($\rho$), reservoir connectivity ($r$), noise in state update ($\xi$), input scaling ($\omega_i$), teacher scaling ($\omega_o$), feedback scaling ($\omega_f$), $\mathrm{L}_2$ ridge regression normalization ($\lambda$). TORNN: processing units per group ($N$), intra-group connections probability ($p$), infra-group connection probability ($q$), spectral radius of $\mathbf{W}_r$ ($\rho$), $\mathrm{L}_2$ regularization of $\mathbf{W}_o$ weights ($\lambda$) and gradient truncation in BPPT ($\tau_{tnc}$).}
  \vspace{-0.05cm}
  \begin{tabular}{r|cccccccc}
    \cmidrule[1.5pt]{1-9}
    \multirow{2}{*}{\textbf{ERNN}} & \multicolumn{2}{c}{$\mathbf{N_r}$} & \multicolumn{2}{c}{\textbf{\# params}} & \multicolumn{2}{c}{$\boldsymbol{\lambda}$} & \multicolumn{2}{c}{$\boldsymbol{\tau_{tnc}}$} \\
     & \multicolumn{2}{c}{91} & \multicolumn{2}{c}{8555} & \multicolumn{2}{c}{1E-5} & \multicolumn{2}{c}{10} \\
    \midrule
    \multirow{2}{*}{\textbf{LSTM}} & \multicolumn{2}{c}{$\mathbf{N_r}$} & \multicolumn{2}{c}{\textbf{\# params}} & \multicolumn{2}{c}{$\boldsymbol{\lambda}$} & \multicolumn{2}{c}{$\boldsymbol{\tau_{tnc}}$} \\
     & \multicolumn{2}{c}{45} & \multicolumn{2}{c}{8506} & \multicolumn{2}{c}{1E-5} & \multicolumn{2}{c}{10} \\
    \midrule
    \multirow{2}{*}{\textbf{GRU}} & \multicolumn{2}{c}{$\mathbf{N_r}$} & \multicolumn{2}{c}{\textbf{\# params}} & \multicolumn{2}{c}{$\boldsymbol{\lambda}$} & \multicolumn{2}{c}{$\boldsymbol{\tau_{tnc}}$} \\
     & \multicolumn{2}{c}{52} & \multicolumn{2}{c}{8477} & \multicolumn{2}{c}{1E-5} & \multicolumn{2}{c}{10} \\
    \midrule
    \multirow{2}{*}{\textbf{ESN}} & $\mathbf{N_r}$ & $\boldsymbol{\rho}$ & $\mathbf{r}$ & $\boldsymbol{\xi}$ & $\boldsymbol{\omega_i}$ & $\boldsymbol{\omega_o}$ & $\boldsymbol{\omega_f}$ & $\boldsymbol{\lambda}$ \\  & $20 \times K$ & $[0.1,1.5]$ & $[0.1,0.5]$ & $[0,0.1]$ & $[0.1,1]$ & $[0.1,1]$ & $[0.1,1]$ & $[0, 0.5]$ \\
    \midrule
    \multirow{2}{*}{\textbf{TORNN}} & $\mathbf{N}$ & $\mathbf{p}$ & $\mathbf{q}$ & $\boldsymbol{\rho}$ & \multicolumn{2}{c}{$\boldsymbol{\lambda}$} & \multicolumn{2}{c}{$\boldsymbol{\tau_{tnc}}$} \\
    & 20 & 0.4 & 0.1 & 0.95 & \multicolumn{2}{c}{1E-5} & \multicolumn{2}{c}{10} \\
    \cmidrule[1.5pt]{1-9}
  \end{tabular}
  \label{tab:config}
  \end{table*}
\egroup

\begin{figure}[!ht]
	\centering
		\includegraphics[width=0.9\columnwidth]{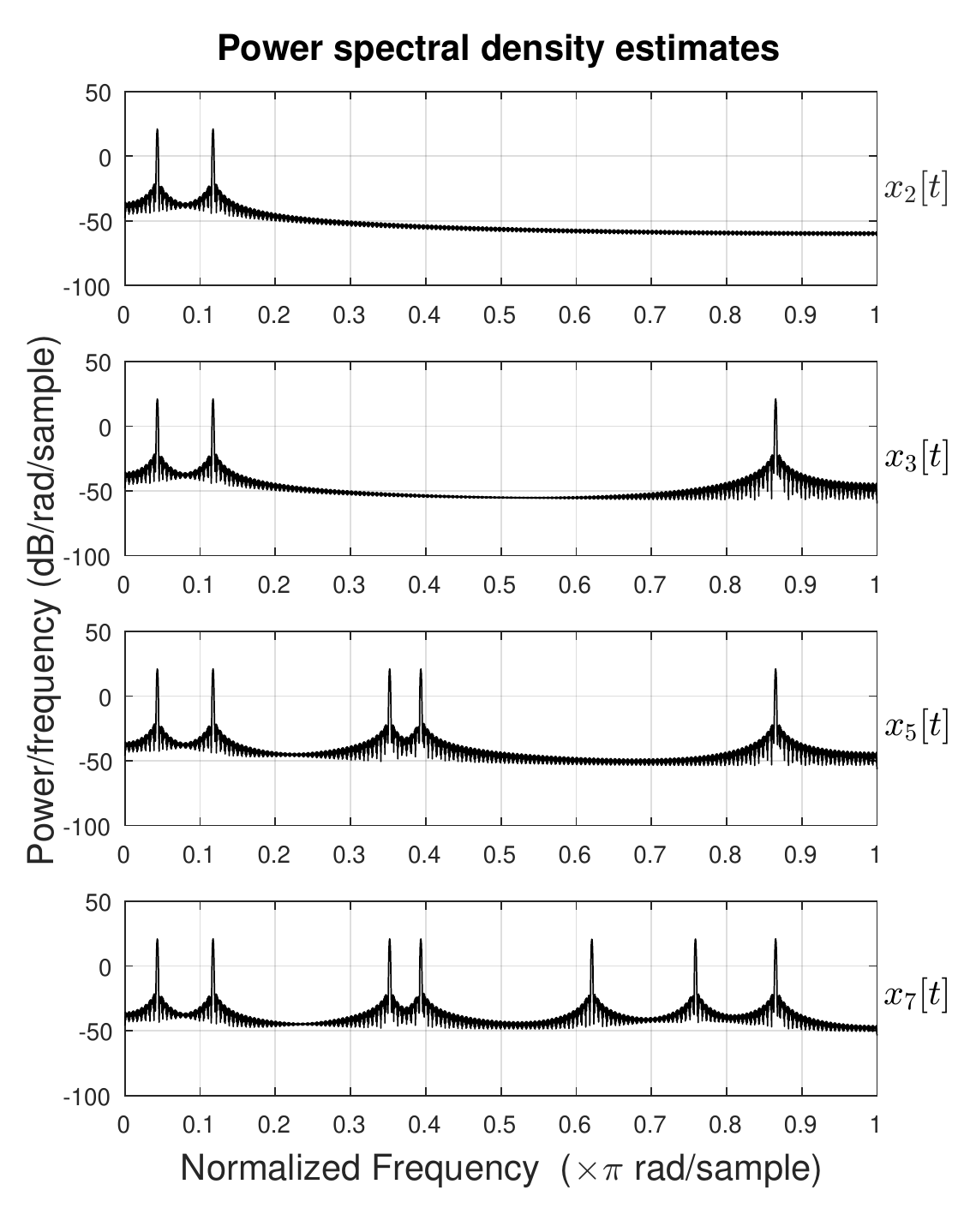}
	\caption{Power spectral densities of the four time series $x_2[t]$, $x_3[t]$, $x_5[t]$ and $x_7[t]$ considered in the experiments. Here, we report the spectrum of the versions without the superimposed white noise.}
	\label{fig:spectra}
\end{figure}

% performance evaluation
In each task we perform a prediction with a forecast horizon of 15 time steps. Prediction error is expressed in terms of Normalized Root Mean Squared Error (NRMSE), computed as
\begin{equation*}
\label{eq:nrmse}
\textrm{NRMSE} = \sqrt{\langle \lVert \boldsymbol{y} - \boldsymbol{y}^* \rVert^2 \rangle/\langle \lVert \boldsymbol{y} - \langle\boldsymbol{y}^*\rangle \rVert^2 \rangle},
\end{equation*}
where $\boldsymbol{y}$ is the output predicted by the network and $\boldsymbol{y}^*$ the ground truth. We consider 4 different time series $x_K[t]$, generated according to Eq. \ref{eq:mso}, with $K=2,3,5,7$ the number of superimposed oscillators. 
The power spectral density estimates of the time series are depicted in Fig. \ref{fig:spectra}. 
As is it possible to see, in the spectra of $x_5[t]$ and $x_7[t]$ there are two frequencies which are very close. 
This increase the difficulty of the problem as these frequencies are harder to separate.
For each time series, we also consider a version with superimposed white noise $n[t]$. 

\begin{figure*}[!ht]
	\centering
		\includegraphics[width=\textwidth]{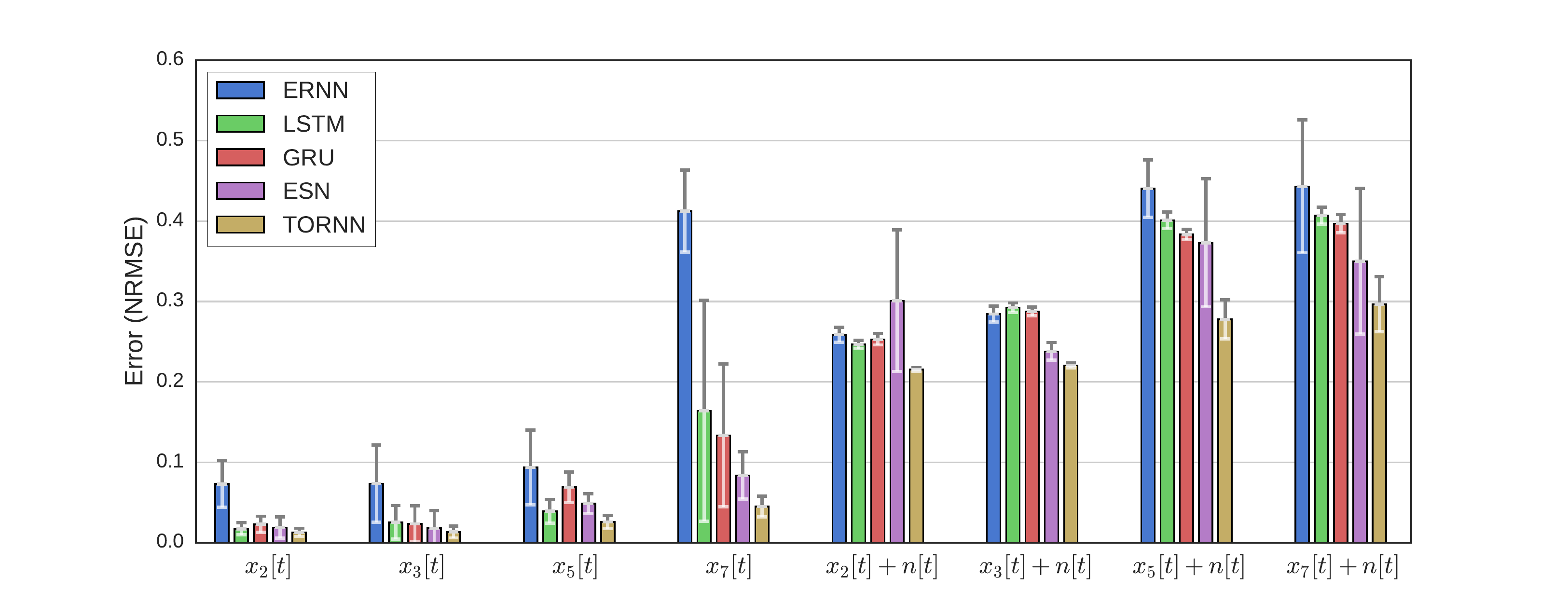}
	\caption{Forecast accuracy (NRMSE) achieved by TORNN and other RNN architectures on the prediction of different time series, relative to a forecast horizon of 15 time intervals. 
	Colored boxes represent the average error on 10 independent trials, the error bars the standard deviation.
	Each time series $x_K[t]$ is the combination of $K$ sinusoids with incommensurable frequencies and they are evaluated with and without a superimposed white noise $n[t]$.}
	\label{fig:results}
\end{figure*}

\begin{figure*}[!ht]
	\centering
		\centerline{\includegraphics[width=0.9\textwidth]{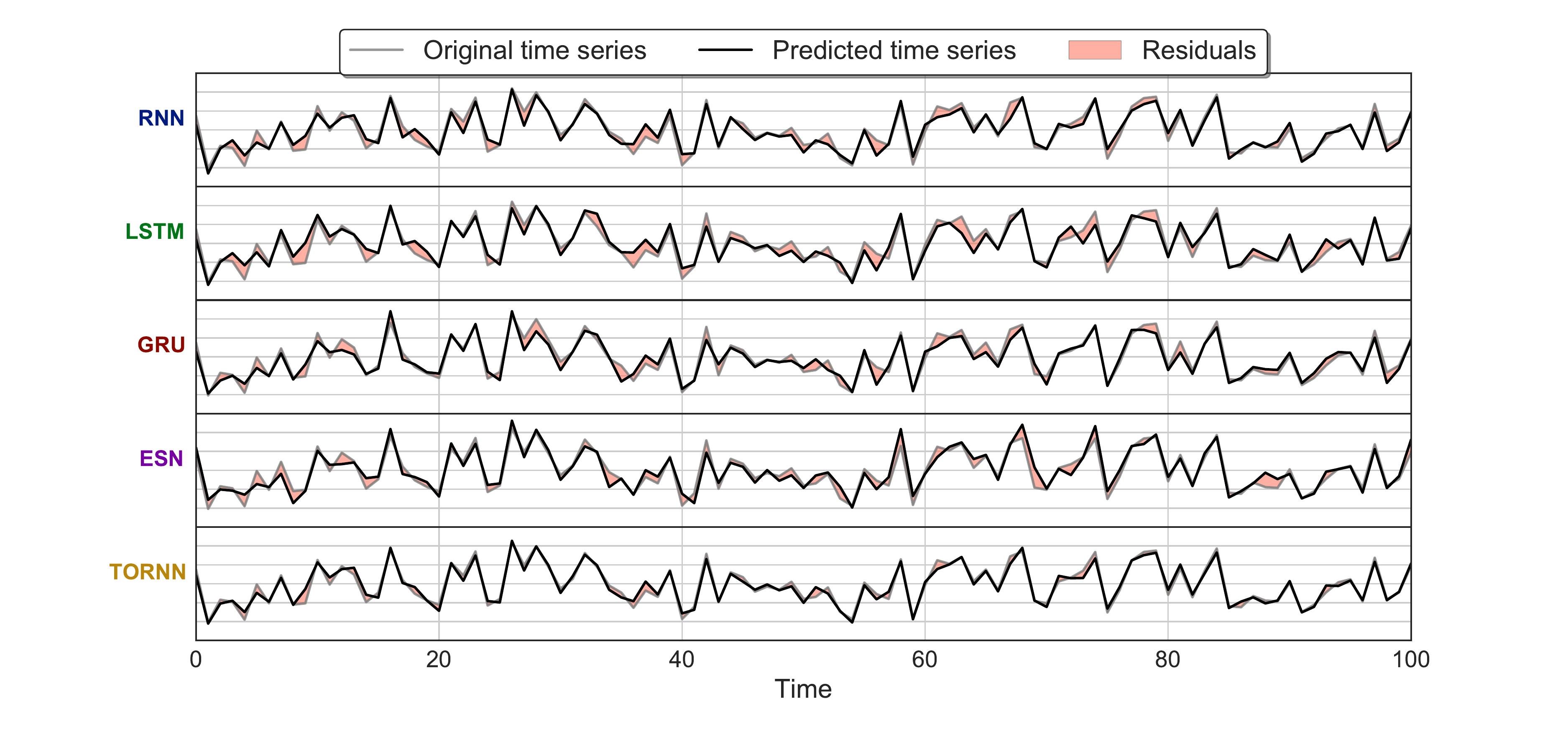}}
	\caption{Prediction results for a fraction of the time series $x_7[t] + n[t]$ with a prediction step-ahead of 15 time steps. For each network, the gray line represents the ground truth, the black line the predicted time series and the shaded red areas are the error residuals of the prediction.}
	\label{fig:plot_pred}
\end{figure*}

The prediction accuracies are reported in Fig. \ref{fig:results}, where the colored boxes represent the average NRMSE error (the shorter, the better) and the error bars are the standard deviations, computed over 10 different trials with independent random initializations. TORNN achieves better performance in every test. 
In order to provide a qualitative description of the forecast results, in Fig. \ref{fig:plot_pred} we show a portion of the prediction of the time series $x_7[t] + n[t]$ by TORNN and the other considered networks, respectively RNN, LSTM, GRU and ESN. In every plot the gray line represents the ground truth, the black line the considered network's forecast and the light red area the residual between the two lines. 
By looking at the residual areas of each time series, it is possible to notice an overall better prediction by TORNN with respect to the other RNNs. 

These results provide empirical evidence of the effectiveness of the proposed framework, which is capable of modeling the different components and thus performing a more accurate prediction.
As expected, when time series are corrupted by noise the prediction forecast accuracy heavily decreases in every model.

% gradient based architectures comparison
The state-of-the-art architectures based on gradient optimization, which are ERNN, LSTM and GRU, obtain on average a larger prediction error with respect to ESN and TORNN. 
Furthermore, ERNN performs worse than LSTM and GRU in every test.
This is a consequence of the simplicity in the architecture, which is unable to handle properly long-term dependencies in the analyzed time series. 
The performance of LSTM and GRU are comparable and it is not possible to conclude which one is the best performing. 
This result is expected and is in agreement with previous studies which evidenced that the selection of a specific gated architecture should depend on the dataset and the corresponding task at hand \cite{DBLP:journals/corr/ChungGCB14}.
However, even if the GRU structure is simpler than the plain LSTM architecture (without peephole connections), the training time of GRU is higher, due to the larger number of operations required to compute the forward and the backward propagation steps \cite{cho2014learning}.

% ESN comparison
The performance of TORNN are always better or at least equal to ESN. 
It is worth noticing that TORNN can be considered as a general case of an ESN, since its architecture reduces to ESN when $\gamma_1 = 1$ and $\gamma_2 = 0$ in every group of units. Indeed, with such a configuration the bandpass filters reduce to regular neurons.
Finally, we underline that in ESN the training of the model is faster than in the other architectures. In fact, rather than a slow gradient descent optimization, the weights of the linear readout in a ESN can be evaluated by performing a simple linear regression.
However, ESNs trade the precision of gradient descent with the ``brute force'' redundancy of random reservoirs and this, inevitably, makes the models more sensitive to the selection of the hyperparameters (like spectral radius). 
Hence, the computational resources required for an accurate search of the optimal hyperparametyers should be accounted in the comparison with other architectures.

%%%%%%%%%%%%%%%%%%%%%%%%%%%%%%%%%%%%%%%%%%%%%%%%%%%%%%%
%%%%%%%%%%%%%%%%%%%%% CONCLUSIONS %%%%%%%%%%%%%%%%%%%%%
%%%%%%%%%%%%%%%%%%%%%%%%%%%%%%%%%%%%%%%%%%%%%%%%%%%%%%%
\section{Conclusions}
\label{sec:conclusions}

In this work we presented the Temporal Overdrive Recurrent Neural Network, a novel RNN architecture designed to model multiple dynamics that are characterized by different time scales.
The proposed model is easy to configure, as it only requires to specify the number of different time scales that should be accounted for, an information that can be easily obtained from a rough frequency analysis. 
Each dynamical component is handled by a group of neurons implementing a bandpass filter, whose behavior is determined by two parameters that determine the cutoff frequencies. 

The proposed methodology follows the strategy of reservoir computing approaches, as the input and recurrent connections are randomly initialized and they are kept untrained. However, while reservoir computing implements a ``brute force'' approach to generate a high, yet redundant, number of internal dynamics, TORNN learns from data the optimal configuration of its dynamics through a gradient descent optimization. Furthermore, with respect to other gradient-based RNNs, the number of trainable parameters is significantly lower, with a consequent simplification of the learning procedure.

We performed some preliminary tests on synthetic data, which showed promising results and demonstrated that our network achieves superior performance in prediction with respect to other state-of-the-art RNNs. 
TORNN can also be used for system identification \cite{nelles2013nonlinear}. In this case, the network must operate in a generative setup and the output of the network has to be fed back into the recurrent layer. 

We are currently working on real-world data, relative to load forecast, which will be presented in a future extension of this work. We also plan to explore the implementation of more efficient bandpass filters, with sharper frequency cutoffs. We believe that this will help in separating the dynamical components more effectively. In the future we also plan to investigate further the internal dynamics of the network.

\bibliographystyle{abbrvnat}
\bibliography{Biblio}
\end{document}